\documentclass[11pt,a4paper]{article}
\usepackage{emnlp2020}
\usepackage{times}
\usepackage{latexsym}
\usepackage{amsmath}
\usepackage{graphicx}
\usepackage{xcolor}
\usepackage{caption}
\usepackage{subfigure}
\usepackage{multirow}
\usepackage{slashbox}
\usepackage{threeparttable}
\usepackage{amssymb}
\usepackage{algorithm}
\usepackage{algorithmic}
\usepackage[algo2e]{algorithm2e}
\usepackage{amsmath}
\usepackage{bbm}

\DeclareMathOperator*{\argmin}{arg\,min}
\definecolor{comment}{rgb}{0.8,0.8,0.8}
\definecolor{applegreen}{rgb}{0.55, 0.71, 0.0}
\usepackage{booktabs}
\usepackage{xcite}

\usepackage{xr}
\makeatletter
\newcommand*{\addFileDependency}[1]{
  \typeout{(#1)}
  \@addtofilelist{#1}
  \IfFileExists{#1}{}{\typeout{No file #1.}}
}
\makeatother

\newcommand{\ours}{{SeqMix}}

\usepackage{microtype}

\aclfinalcopy 
\def\aclpaperid{1720} 

\newif\ifsubmit
\submittrue

\ifsubmit
\newcommand{\chao}[1]{}
\newcommand{\yue}[1]{}
\newcommand{\zc}[1]{}

\else
\newcommand{\chao}[1]{{\color{red}[Chao: #1]}}
\newcommand{\zc}[1]{{\color{red}[Chao: #1]}}
\newcommand{\yue}[1]{{\color{blue}[Yue: #1]}}

\fi

\title{\ours: Augmenting Active Sequence Labeling via Sequence Mixup}

\author{Rongzhi Zhang \\
  Georgia Tech \\
  \texttt{rongzhi.zhang@gatech.edu} \ \ ~~ \\\And
  Yue Yu \\
  Georgia Tech \\
  \texttt{yueyu@gatech.edu}  \\\And
 Chao Zhang \\
  Georgia Tech \\
  \texttt{chaozhang@gatech.edu} \\}

\date{}
\begin{document}
\maketitle

\begin{abstract}

  Active learning is an important technique for low-resource sequence labeling tasks. However,
current active sequence labeling methods
use the queried samples alone in each iteration, which is an inefficient way of leveraging human annotations. We propose a simple but effective data
augmentation method to improve label efficiency of active sequence labeling. Our method, \ours, simply augments the queried samples by generating
extra labeled sequences in each iteration. The key difficulty is to generate plausible sequences along with token-level labels. In {\ours}, we address
this challenge by performing mixup for both sequences and token-level labels of the queried samples. Furthermore, we design a discriminator during
sequence mixup, which judges whether the generated sequences are plausible or not. Our experiments on Named Entity Recognition  and  Event Detection
tasks show that {\ours} can improve the standard active sequence labeling method by $2.27\%$--$3.75\%$ in terms of $F_1$ scores. The code and data for \ours \ can be found at {\url{https://github.com/rz-zhang/SeqMix}}.







\end{abstract}


\section{Introduction}


Many NLP tasks can be formulated as sequence labeling problems, such as part-of-speech (POS) tagging~\cite{zheng-etal-2013-deep}, named entity
recognition (NER)~\cite{lample2016neural}, and event extraction~\cite{yang2019exploring}. Recently, neural sequential models \cite{lample2016neural,akbik2018contextual,transformer} have shown strong performance for various sequence labeling task. However, these deep neural models are \emph{label
  hungry}---they require large amounts of annotated sequences to achieve strong performance. Obtaining large amounts of annotated data can be too
expensive for practical sequence labeling tasks, due to token-level annotation efforts.


Active learning is an important technique for sequence labeling in low-resource settings. Active sequence labeling is an iterative process. In each
iteration, a fixed number of unlabeled sequences are selected by a query policy for annotation and then model updating, in hope of
maximally improving model performance. For example,~\citet{tomanek2007approach,shen2017deep} select query samples based on data uncertainties; ~\citet{hazra2019active} compute model-aware similarity to eliminate redundant examples and improve the diversity of query samples; and \citet{fang2017learning,liu2018learning} use reinforcement learning to learn query policies. However, existing methods for active sequence labeling all use the queried samples \textit{alone} in each iteration. We argue that the queried samples provide limited data diversity, and using them alone for model updating is inefficient in terms of leveraging human annotation efforts.

We study the problem of enhancing active sequence labeling via data
augmentation. We aim to generate augmented labeled sequences for the queried
samples in each iteration, thereby introducing more data diversity and improve
model generalization. However, data augmentation for active sequence labeling is
challenging, because we need to  generate sentences and token-level
labels jointly. Prevailing generative models
\cite{zhang2016generating,bowman2015generating} are inapplicable because they
can only generate word sequences without labels. It is also infeasible to apply
heuristic data augmentation methods such as context-based words substitution \cite{kobayashi2018contextual},
synonym replacement, random insertion, swap, and deletion
\cite{wei2019eda}, paraphrasing~\cite{paraphrase} or back translation~\cite{xie2019unsupervised}, because label composition is complex
for sequence labeling. Directly using these techniques to manipulate tokens may
inject incorrectly labeled sequences into training data and harm model
performance.


We propose \ours, a data augmentation method for
generating sub-sequences along with
their labels
based
on \emph{mixup} \cite{zhang2017mixup}. Under the active sequence labeling framework, {\ours} is capable of
generating plausible pseudo labeled sequences for the queried samples in each
iteration. This is enabled by two key techniques in {\ours}: (1) First, in each
iteration, it searches for pairs of eligible
sequences and mixes them both in the feature space and the label space. (2)
Second, it has a discriminator to judge if the generated sequence is plausible
or not. The discriminator is designed to compute the perplexity scores for all
the generated candidate sequences and select the low-perplexity sequences as
plausible ones.



We show that {\ours}
consistently outperforms standard active sequence labeling
baselines under different data usage percentiles
with
experiments on Named Entity Recognition and Event Detection tasks. On average, it achieves $2.95\%, 2.27\%, 3.75\%$
$F_1$ improvements on the CoNLL-2003, ACE05 and WebPage datasets. 
The advantage of {\ours} is especially prominent in low-resource scenarios, achieving $12.06\%, 8.86\%, 16.49\%$ $F_1$ improvements to the original active learning approach on the above three datasets.  
Our results also verify the proposed mixup strategies and the discriminator are vital to
the performance of \ours.

\section{Preliminaries}


\subsection{Problem Definition}
\label{prob_setting}

Many NLP problems can be formulated as sequence labeling problems. Given an
input sequence, the task is to annotate it with token-level labels. The labels
often consist of a position prefix provided by a labeling schema and a type indicator
provided by the specific task. For example, in the named entity recognition
task, we can adopt the \texttt{BIO} (Beginning, Inside, Outside) tagging
scheme~\cite{marquez2005semantic} to assign labels for each token: the first
token of an entity mention with type \texttt{X} is labeled as \texttt{B-X}, the
tokens inside that mention are labeled as \texttt{I-X} and the non-entity
tokens are labeled as \texttt{O}.

 Consider a large unlabeled corpus $\mathcal{U}$, traditional active learning starts from a small annotated seed set $\mathcal{L}$, and utilizes a query function $\psi(\mathcal{U}, K, \gamma(\cdot))$ to obtain $K$ most informative unlabeled samples $\mathcal{X}=\left\{\mathbf{x}_{1}, \ldots, \mathbf{x}_{K}\right\}$ along with their labels $\mathcal{Y}=\left\{y_{1}, \cdots, y_{K}\right\}$, where $\gamma(\cdot)$ is the query policy.
Then, we remove $\mathcal{X}$ from the unlabeled data $\mathcal{U}$ and 
repeat the above procedure until the satisfactory performance achieved or the annotation capacity reached.

In \ours, we aim to further exploit the annotated set $ \left\langle\mathcal{X}, \mathcal{Y}
\right\rangle$ to generate augmented data
$ \left\langle\mathcal{X}^*, \mathcal{Y}^* \right\rangle$.
Then the labeled dataset is expanded as $\mathcal{L} = \mathcal{L} \cup \left\langle\mathcal{X}, \mathcal{Y} \right\rangle \cup  \left\langle\mathcal{X}^*, \mathcal{Y}^* \right\rangle$.
Formally, we define our task as:
(1) construct a generator $\phi(\cdot)$ to implement sequence and label generation based on the actively sampled data $\mathcal{X}$ and its label $\mathcal{Y}$, (2) set a discriminator $d(\cdot)$ to yield the filtered generation, then (3) augment the labeled set as
$\mathcal{L} = \mathcal{L} \cup \left\langle \mathcal{X}, \mathcal{Y}\right\rangle \cup d(\phi( \mathcal{X}, \mathcal{Y}))$.

\subsection{Active Learning for Sequence Labeling}
\label{sampling_policy}


Active sequence labeling selects $K$ most informative instances $\psi\left(\cdot, K,
\gamma(\cdot)\right)$ in each iteration, with the hope of maximally improving model performance
with a fixed labeled budget. With the input sequence $\mathbf{x}$ of length $T$,
we denote the model output as $f\left(\cdot | \mathbf{x} ; \theta\right)$.
Our method is
 generic to any query policies $\gamma(\cdot)$.
 Below, we introduce several representative policies.

\paragraph{Least Confidence~(LC)}
\label{lc}
\citet{culotta2005reducing} measure the uncertainty of sequence models by
the most likely predicted sequence. For a CRF model~\cite{lafferty2001conditional}, we calculate $\gamma$ with the predicted sequential label $\mathbf{y}^*$ as
\begin{equation}
       \label{eq:lc_1}
    \gamma^{\text{LC}}(\mathbf{x})=1- \max_{y^{*}}(P\left(\mathbf{y}^{*} | \mathbf{x}; \theta\right),
\end{equation}
where $\mathbf{y}^{*}$ is the Viterbi parse. 
For BERT~\cite{devlin2018bert} with a token classification head, we adopt a variant of the least confidence measure:
\begin{equation}
\label{eq:lc_2}
    \gamma^{\text{LC'}}(\mathbf{x})=\sum_{t=1}^T(1- \max_{\mathbf{y}_t}P\left(\mathbf{y}_t | \mathbf{x} ; \theta\right)),
\end{equation}
where $P(\mathbf{y}_t | \mathbf{x}; \theta)  = \text{softmax}(f(\mathbf{y}_t | \mathbf{x}; \theta))$.

\paragraph{Normalized Token Entropy (NTE)}
\label{nte}
Another uncertainty measure for the query policy is normalized entropy
\cite{settles2008analysis}, defined as:
\begin{equation}
\label{eq:nte}
    \gamma^{\text{TE}}(\mathbf{x})=-\frac{1}{T} \sum_{t=1}^{T} \sum_{m=1}^{M} P_m(\mathbf{y}_{t}|\mathbf{x},{\theta}) \log P_m(\mathbf{y}_{t}|\mathbf{x},{\theta}),
\end{equation}
where $P_m(\mathbf{y}_{t}|\mathbf{x},{\theta}) = [\text{softmax}(f(\mathbf{y}_t | \mathbf{x}; \theta))]_m$. 

\paragraph{Disagreement Sampling}
\label{disagreement}
Query-by-committee (QBC)  \cite{seung1992query}, 
is another approach for specifying the policy,
where the unlabeled data can be sampled by the disagreement of the base models. The disagreement can be defined in several ways, here we take the vote entropy proposed by \cite{dagan1995committee}.
Given a committee consist of $C$ models, the vote entropy for input $\mathbf{x}$ is:
\begin{equation}
\label{eq:disagree}
    \gamma^{\text{VE}}(\mathbf{x})=-\frac{1}{T} \sum_{t=1}^{T} \sum_{m=1}^{M} \frac{V_m\left(\mathbf{y}_{t}\right)}{C} \log \frac{V_m\left(\mathbf{y}_{t}\right)}{C},
\end{equation}
where $V_m(\mathbf{y}_{t})$ is the number of models that predict the $t$-th token $\mathbf{x}_t$ as the label $m$.


\section{The \ours \ Method}



\subsection{Overview}


Given a corpus
for sequence labeling, we assume the dataset  contains
a small labeled set $\mathcal{L}$ and a large unlabeled set $\mathcal{U}$ initially.
We start from augmenting the seed set $\mathcal{L}$ with \ours. First, we adopt a pairing function $\zeta(\cdot)$  to find paired samples by traversing  $\mathcal{L}$. Next, we generate mixed-labeled sequences via latent space linear interpolation with one of the approaches mentioned in Section \ref{mixup_emb}.
To ensure the semantic quality of the generated sequences, we use a discriminator $d(\cdot)$ to measure the perplexity of them and filter low-quality sequences out. Then we generate the extra labeled sequences $\mathcal{L}^* = \textit{SeqMix}(\mathcal{L}, \alpha, \zeta(\cdot), d(\cdot))$ and get the augmented training set $ \mathcal{L} = \mathcal{L} \cup \mathcal{L}^*$. The sequence labeling model $\theta$ is initialized on this augmented training set $\mathcal{L}$.

After that, the iterative active learning procedure begins. In each iteration, we actively select instances from $\mathcal{U}$ with a query policy $\gamma(\cdot)$ (Section \ref{sampling_policy}) to obtain the top $K$ samples $\mathcal{X} = \psi(\mathcal{U}, K, \gamma(\cdot))$. The newly selected samples will
 be labeled with $\mathcal{Y}$, and the batch of samples $\left\langle \mathcal{X}, \mathcal{Y} \right\rangle$ will be used for \ours. Again, we
 generate $\mathcal{L}^* = \textit{SeqMix}(\left\langle \mathcal{X}, \mathcal{Y} \right\rangle, \alpha, \zeta(\cdot), d(\cdot))$ and expand the
 training set as  $\mathcal{L} = \mathcal{L} \cup \mathcal{L}^*$. Then we train the model $\theta$ on the newly augmented set $\mathcal{L}$. The iterative
 active learning procedure terminates when a fixed number of iterations are reached. We summarize the above procedure in Algorithm \ref{algo:augmentation}.

\begin{algorithm}[h]
\SetAlgoLined
\textbf{Input}:
Labeled seed set $\mathcal{L}$;
Unlabeled set $\mathcal{U}$;
Query function $\psi(\cdot, K, \gamma(\cdot))$;
The sequence labeling model $\theta$;
Beta distribution parameter $\alpha$;
Pairing function $\zeta(\cdot)$; Discriminator function $d(\cdot)$.\\
\textbf{// seed set augmentation}
\begin{center}
    $\mathcal{L}^* = \textit{SeqMix}(\mathcal{L}, \alpha, \zeta(\cdot), d(\cdot))$\\
 $\mathcal{L} = \mathcal{L} \cup \mathcal{L}^*$
\end{center}
 \textbf{// model initialization}
 \begin{center}
     $\theta$ = train ($\theta, \mathcal{L}$)
 \end{center}
 \textbf{// active learning iterations with augmentation}\\
 \For{round \textbf{in} active learning rounds}
 {
 $\mathcal{X} = \psi(\mathcal{U}, K, \gamma(\cdot))$\\
  $\mathcal{U} = \mathcal{U} - \mathcal{X}$\\
 Annotate $\mathcal{X}$ to get $\left\langle \mathcal{X}, \mathcal{Y} \right\rangle$\\
 $\mathcal{L}^* = \textit{SeqMix}(\left\langle \mathcal{X}, \mathcal{Y} \right\rangle, \alpha, \zeta(\cdot), d(\cdot))$\\
  $\mathcal{L} = \mathcal{L} \cup \left\langle \mathcal{X}, \mathcal{Y} \right\rangle \cup \mathcal{L}^*$\\
 $\theta$ = train ($\theta, \mathcal{L}$)
 }
 \textbf{Output}:
The sequence model trained with active data augmentation: $\theta$
 \caption{The procedure of active sequence labeling augmentation via \ours}
\label{algo:augmentation}
\end{algorithm}
\subsection{Sequence Mixup in the Embedding Space}
\label{mixup_emb}
Mixup \cite{zhang2017mixup} is a data augmentation method that implements linear interpolation in the input space. Given two input samples $x_{i}, x_{j}$ along with the labels $y_{i}, y_{j}$, the mixing process is:
\begin{align}
\label{mixup}
    &\tilde{x}=\lambda x_{i}+(1-\lambda) x_{j},\\
    &\tilde{y}=\lambda y_{i}+(1-\lambda) y_{j},
\end{align}
 where $\lambda \sim \textit{Beta}(\alpha, \alpha)$ is the mixing coefficient. Through linear combinations on the input level of paired examples and their labels, Mixup regularizes the model to present linear behavior among the training data.

Mixup is not directly applicable to generate interpolated samples for text data, because the input space is discrete. To overcome this, SeqMix performs token-level
interpolation in the embedding space and selects a token closest to the interpolated embedding. Specifically, {\ours} constructs a table of tokens $\mathcal{W}$ and their corresponding contextual embeddings $\mathcal{E}$\footnote{The construction of $\{\mathcal{W},\mathcal{E}\}$ are discussed in Appendix.}. Given two sequences $\mathbf{x_i} = \{\mathbf{w}^1_i,\cdots, \mathbf{w}^T_i\}$ and $\mathbf{x_j} = \{\mathbf{w}^1_j,\cdots, \mathbf{w}^T_j\}$
with their embedding representations
$\mathbf{e_{x_i}} =  \{ \mathbf{e}^1_i,\cdots, \mathbf{e}^T_i\}$ and $\mathbf{e_{x_j}} =  \{ \mathbf{e}^1_j,\cdots, \mathbf{e}^T_j\}$, the $t$-th mixed token is the token whose embedding $\mathbf{e}^{t}$ is closest to the mixed embedding:
\begin{equation}
\label{emb_mix}
    \mathbf{e}^{t} = \argmin_{\mathbf{e} \in \mathcal{E}} \left\| \mathbf{e} -
    (\lambda \mathbf{e_{i}^t}+(1-\lambda) \mathbf{e_{j}^t})\right\|_2.
\end{equation}
To get the corresponding $\mathbf{w}^t$, we can query the table $\{\mathcal{W}, \mathcal{E}\}$ using  $\mathbf{e}^{t}$.
The label generation is straightforward. For two label sequences $\mathbf{y}_i = \{\mathbf{y}^1_i,\cdots, \mathbf{y}^T_i\}$ and $\mathbf{y}_j = \{\mathbf{y}^1_j,\cdots, \mathbf{y}^T_j\}$, we get the $t$-th mixed label as:
\begin{equation}
\label{label_mix}
    \mathbf{y}^t=\lambda \mathbf{y}^t_i+(1-\lambda) \mathbf{y}^t_j,
\end{equation}
where $\mathbf{y}^t_i$ and  $\mathbf{y}^t_j$ are one-hot encoded labels.

Along with the above sequence mixup procedures, we also introduce a pairing strategy that selects sequences for mixup. The
reason is that, in many sequence labeling tasks, the labels of interest are scarce. For example, in the NER and event detection tasks, the ``O'' label
is dominant in the corpus, which do not refer to any entities or events of interest. We thus define the labels of interest as \textit{valid labels},
\textit{e.g.}, the non-``O'' labels in NER and event detection, and design a sequence pairing function to select more informative parent sequences for mixup.
Specifically, the sequence pairing function $\zeta(\cdot)$ is designed according to \textit{valid label density}. For a sequence, its valid label
density is defined as $\eta = \frac{n}{s}$, where $n$ is the number of valid labels and $s$ is the length of the sub-sequence. We set a threshold
$\eta_0$ for $\zeta(\cdot)$,  and the sequence will be considered as an eligible candidate for mixup only when $\eta \geq \eta_0$.

Based on the above token-level mixup procedure and the sequence pairing function, we propose three different
strategies for generating interpolated labeled sequences. These strategies are shown in
Figure \ref{fig:3mixup} and described below:

\begin{figure*}[!h]
    \centering
    \subfigure[Whole sequence mixup]
    {\includegraphics[scale=0.15]{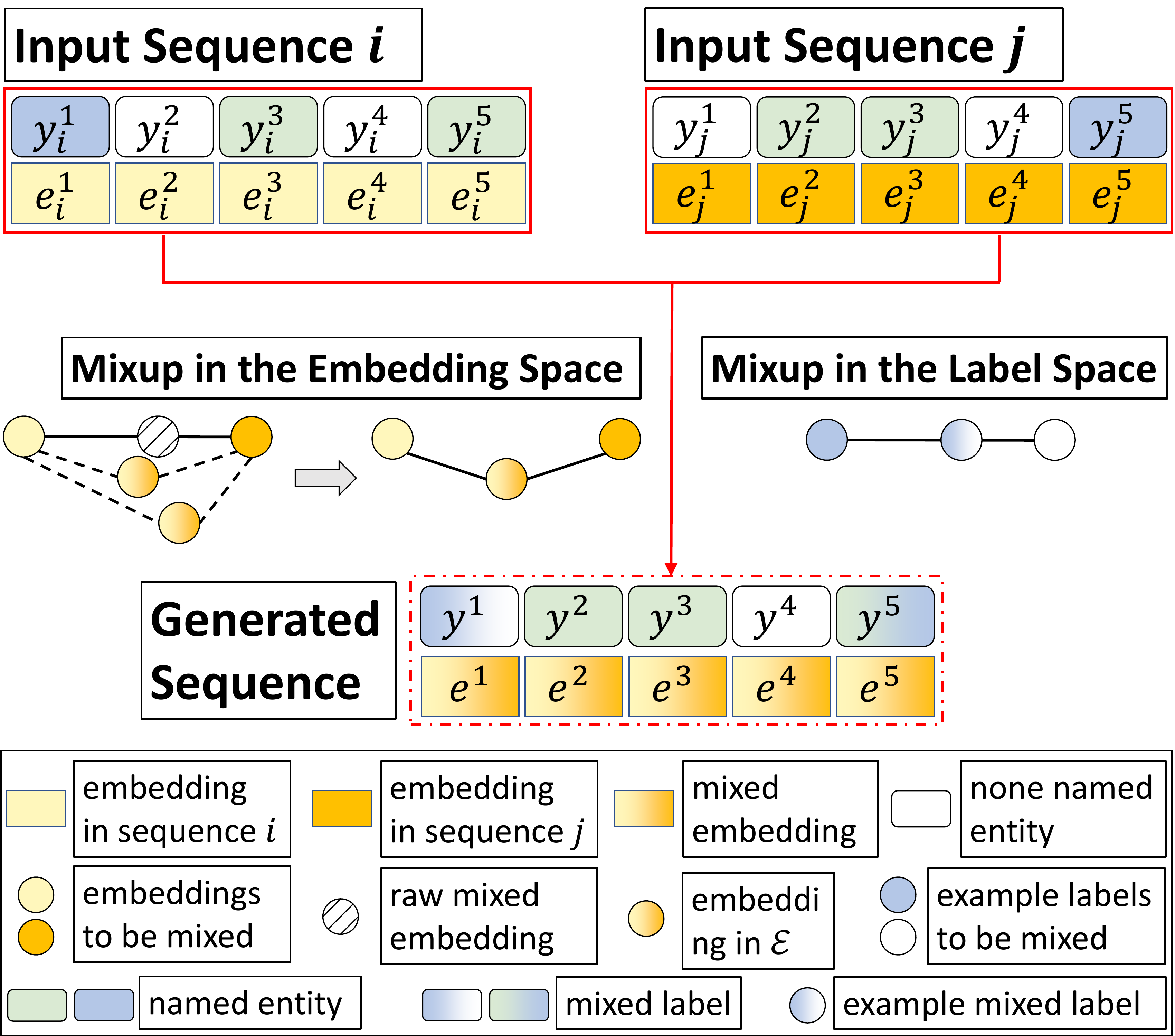}
        \label{fig:wholemixup}}
    \subfigure[Sub-sequence mixup]       
    {\includegraphics[scale=0.15]{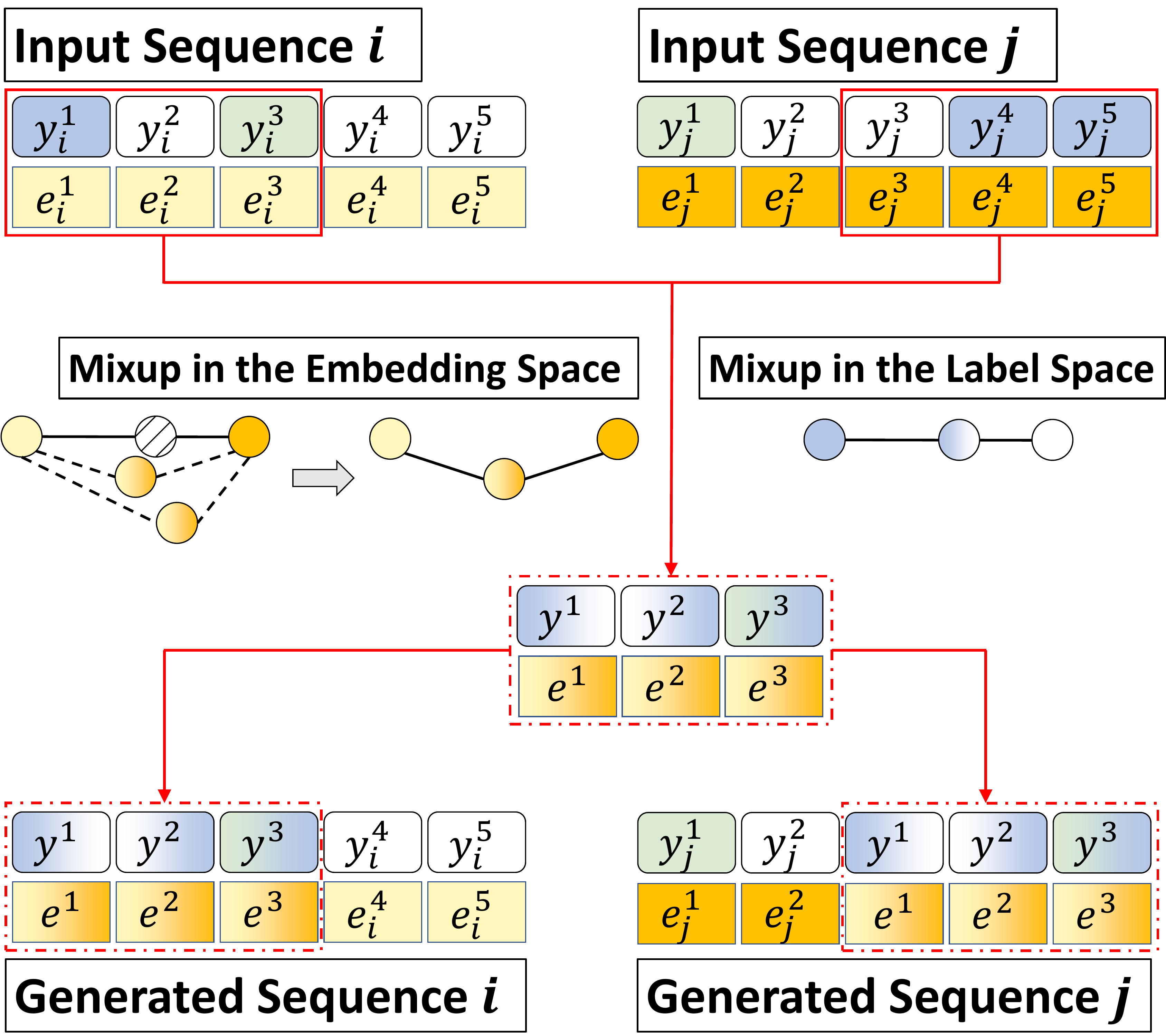}
        \label{fig:submixup} }
    \subfigure[Label-constrained sub-sequence mixup] {\includegraphics[scale=0.15]{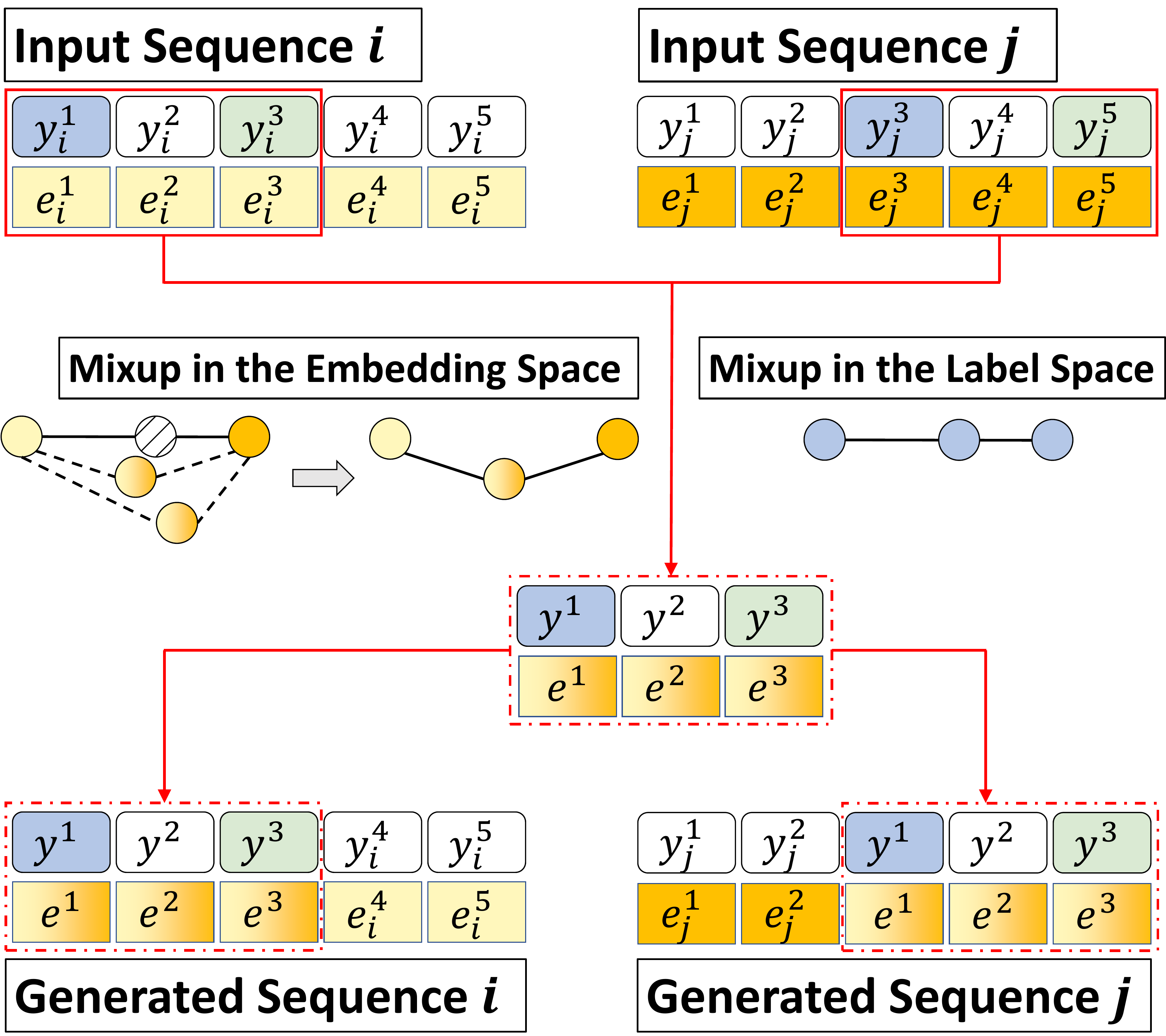}
        \label{fig:lcmixup}}
    \caption{Illustration of the three variants of {\ours}. We use $s=5, \eta_0 = \frac{3}{5}$ for whole-sequence mixup and $s=3, \eta_0 = \frac{2}{3}$ for sub-sequence mixup and label-constrained sub-sequence mixup. The solid red frames indicate paired sequences or sub-sequences, and the red dotted frames indicate generated sequence or sub-sequence. In the original sequences, the parts not included in the solid red frames will be unchanged in the generated sequences. For the mixup in the embedding space, we take the embedding in $\mathcal{E}$ which is closest to the raw mixed embedding as the generated embedding. For the mixup in the label space, the mixed label can be used as the pseudo label. }
    \label{fig:3mixup}
\end{figure*}

\begin{algorithm}[ht]
	\DontPrintSemicolon
\textbf{Input}:
Labeled set $ \mathcal{L} = \left\langle \mathcal{X}, \mathcal{Y} \right\rangle$;
Beta distribution parameter $\alpha$;
Pairing function $\zeta(\cdot)$; Discriminator function $d(\cdot)$; Number of expected generation $N$.\\
\For {$\left\langle \mathbf{x_i, y_i} \right\rangle, \left\langle \mathbf{x_j, y_j} \right\rangle, (i\neq j) $ in $\mathcal{L}$}
{
    \If{$\zeta(\left\langle \mathbf{x_i, y_i} \right\rangle, \left\langle \mathbf{x_j, y_j} \right\rangle)$}
    {
        $\lambda \sim$ \textit{Beta}$(\alpha, \alpha)$ \\
        // \textbf{mixup the target sub-sequences} \\
        \For{$t = 1,\cdots, T$}
        {
            Calculate $\mathbf{e}^t$  by Eq.~\eqref{emb_mix};\\
            Get corresponding token $\mathbf{w}^t$ \text{for} $\mathbf{e}^t$;  \\
            Calculate $\mathbf{y}^t$  by Eq.~\eqref{label_mix}.
        }
        $\tilde{\mathbf{x}}_{\textit{sub}}= \{\mathbf{w}^1, \cdots, \mathbf{w}^T\}$\\
        $\tilde{\mathbf{y}}_{\textit{sub}}= \{\mathbf{y}^1, \cdots, \mathbf{y}^T\}$\\
        // \textbf{replace the original sequences}\\
        \For{$k$ in $\left\{ i, j\right\}$}
        {
             $\tilde{\mathbf{x}}_{k} = \mathbf{x}_k - \mathbf{x_{k\textit{sub}}} + \tilde{\mathbf{x}}_{\textit{sub}} $\\
              $\tilde{\mathbf{y}}_{k} = \mathbf{y}_k - \mathbf{y_{k\textit{sub}}} + \tilde{\mathbf{y}}_{\textit{sub}} $\\
              \If{$d(\tilde{\mathbf{x_k}})$}
                {
                    $\mathcal{L}^* = \mathcal{L}^* \cup \left\langle \tilde{\mathbf{x}}_k, \tilde{\mathbf{y}}_k \right\rangle$
                }
              \If{$|\mathcal{L}^{*}| \geq N $}
              {break}
        }
    }
}
\KwOut{Generated sequences and labels $\mathcal{L}^{*}$}
\caption{The generation procedure of \ours}
\label{algo:seqmix}
\end{algorithm}

\paragraph{Whole-sequence mixup}
\label{softmix}
As the name suggests, whole-sequence mixup (Figure \ref{fig:wholemixup}) performs sequence mixing at the whole-sequence level.
Given two sequences $\left\langle \mathbf{x_i}, \mathbf{y_i} \right\rangle, \left\langle \mathbf{x_j}, \mathbf{y_j} \right\rangle \in \mathcal{L}$,
they must share the same length without counting padding words.
Besides, the paring function $\zeta(\cdot)$ requires that both the two sequences satisfy $\eta \geq \eta_0$. 
Then we perform mixup at all token positions, by employing
Equation \ref{emb_mix} to generate mixed tokens and
Equation \ref{label_mix} to generate mixed labels (note that the mixed labels are soft labels).

\paragraph{Sub-sequence mixup}
\label{slackmix}
One drawback of the whole-sequence mixup is that it
indiscriminately mixes over all tokens, which
may include incompatible subsequences and generate implausible sequences.
To tackle this, we consider sub-sequence mixup (Figure \ref{fig:submixup}) 
to mix sub-sequences of
the parent sequences. It scans the original samples with a window of fixed-length $s$ to look for paired sub-sequences.
Denote the sub-sequences of $\left\langle \mathbf{x_i}, \mathbf{y_i} \right\rangle, \left\langle \mathbf{x_j}, \mathbf{y_j} \right\rangle $ as $\mathbf{X_{i\textit{sub}}}=\left\{\mathbf{x}^1_{\mathbf{i}\textit{sub}}, \ldots, \mathbf{x}^s_{\mathbf{i}\textit{sub}}\right\}$,
$\mathbf{X_{j\textit{sub}}}=\left\{\mathbf{x}^1_{\mathbf{j}\textit{sub}}, \ldots, \mathbf{x}^s_{\mathbf{j}\textit{sub}}\right\}$.
If $\exists$ $\mathbf{x_{i\textit{sub}}} \in \mathbf{X_{i\textit{sub}}}$,
$\mathbf{x_{j\textit{sub}}} \in \mathbf{X_{j\textit{sub}}}$,
such that their $\eta \geq \eta_0$, we have $\zeta(\left\langle \mathbf{x_i}, \mathbf{y_i} \right\rangle, \left\langle \mathbf{x_j}, \mathbf{y_j} \right\rangle) = \text{True}$. Then the sub-sequences $\mathbf{x_{i\textit{sub}}}$ and $\mathbf{x_{j\textit{sub}}}$ are mixed as Figure \ref{fig:submixup}. The mixed sub-sequence and labels will replace the original parts of the parents samples, and the other parts of the parent samples remain unchanged. In this way, sub-sequence mixup is expected to keep the syntax structure of the original sequence, while providing data diversity.

\paragraph{Label-constrained sub-sequence mixup}
\label{lfmix}
 can be considered as a special case of sub-sequence mixup, where the constraints inherit sub-sequence mixup, and further require that the
 sub-sequence labels are consistent. As Figure \ref{fig:lcmixup} shows, after mixing such paired samples, the generation will just update the tokens of the sub-sequences while keeping the labels the same as before. Hence, this version is called label-constrained sub-sequence mixup.

Comparing the three variants, label-constrained sub-sequence mixup gives the most restrictions to pairing parent samples, sub-sequence mixup sets the sub-sequence-level pattern, while whole-sequence mixup just requires $\eta \geq \eta_0$ for the sequences with the same length.




\subsection{Scoring and Selecting Plausible Sequences}
\label{scoreing}


During sequence mixup, the mixing coefficient $\lambda$ determines the strength of interpolation. When $\lambda$ approximates 0 or 1, the generated
sequence will be similar to one of the parent sequences, while the $\lambda$ around $0.5$ produces relatively diverse generation. However, generating
diverse sequences means low-quality sequences can be generated, which can provide noisy contextual information and  hurt model performance.


To maintain the quality of mixed sequences, we set a discriminator to score the perplexity of the sequences. The final generated sequences will consist of only the sequences that pass the sequence quality screening. For
screening, we utilize a language model GPT-2 \cite{radford2019language} to score sequence $\mathbf{x}$ by computing its perplexity:
\begin{align}
\label{ppl}
    &\rm{Perplexity}(\mathbf{x}) = 2^{-\frac{1}{T} \sum_{i=1}^{T} \log p\left(w_{i}\right)},
\end{align}
where $T$ is the number of tokens before padding, $w_i$ is the $i$\textit{-th} token of sequence $\mathbf{x}$. Based on the perplexity and a score range $[s_1, s_2]$, the discriminator can give judgment for sequence $\mathbf{x}$:
\begin{equation}
    d(\mathbf{x}) =
    \mathbbm{1}\left\{s_1 \leq \rm{Perplexity}\left(\mathbf{x}\right) \leq s_2\right\}.
\end{equation}

The lower the perplexity score, the more natural the sequence. However, the discriminator should also consider the regularization effectiveness and the generation capacity. Hence, a blind low perplexity setting is undesirable.
The overall sequence mixup and selection procedure is illustrated in Algorithm \ref{algo:seqmix}.


\section{Experiments}
\subsection{Experiment Setup}
\label{exp_setup}
\textbf{Datasets.} \
We conduct experiments on three sequence labeling datasets for the named entity recognition (NER)  and
event detection tasks. 

\noindent (1) \textbf{CoNLL-03} \cite{sang2003introduction} is a corpus for NER task. It provides four named entity types: persons, locations, organizations, and miscellaneous.\footnote{We take the English version as our target corpus.}

\noindent (2) \textbf{ACE05} is a corpus for 
event detection.  It provides 8 event types and 33 subtypes.  We study the  event trigger detection problem, which aims to identify trigger tokens in
a sentence. 

\noindent (3) \textbf{Webpage} \cite{ratinov2009design} is a NER corpus with 20 webpages related to computer science conference and academic websites. It inherits the entity types from CoNLL-03.



\noindent \textbf{Data Split.} \  To investigate low-resource sequence labeling, we randomly take 700 labeled sentences from the original CoNLL-03 dataset as the training set. For ACE05 and WebPage dataset, the annotation is sparse, so we conduct experiments on their original dataset without further slicing. 

 We set 6 data usage percentiles for the training set in each corpus. The sequence model is initialed on a small seed set, then it performs five
 iterates of active learning. For the query policy, we use random sampling and the three active learning policies mentioned in Section
 \ref{sampling_policy}. The machine learning  performance is evaluated by $F_1$ score for each data usage percentile.

\noindent \textbf{Parameters.} \ We use BERT-base-cased for the NER task as the underlying model, and BERT-base-multilingual-cased for the event
trigger detection task. We set the max length as 128 to pad the varying-length sequences. The learning rate of the underlying model is 5e-5, and the
batch size is 32. We trained them for 10 Epochs at each data usage percentile. For the parameters of {\ours}, we set the $\alpha = 8$ to sample
$\lambda$ from $\textit{Beta} (\alpha, \alpha)$. We use the sub-sequence window length $s = \{5, 5, 4\}$, the valid label density $\eta_0 = \{0.6,
0.2, 0.5\}$ for CoNLL-03, ACE05 and Webpage, respectively. The augment rate is set as 0.2, and the discriminator score range is set as $(0, 500)$. We
also perform a detailed parameter study in Section \ref{param_study}.

\subsection{Results}


\begin{figure*}[t]
    \vspace{-0.05in}
        \centering
        \subfigure[CoNLL-03 (700 labeled data)]{
            \includegraphics[width=0.322\textwidth]{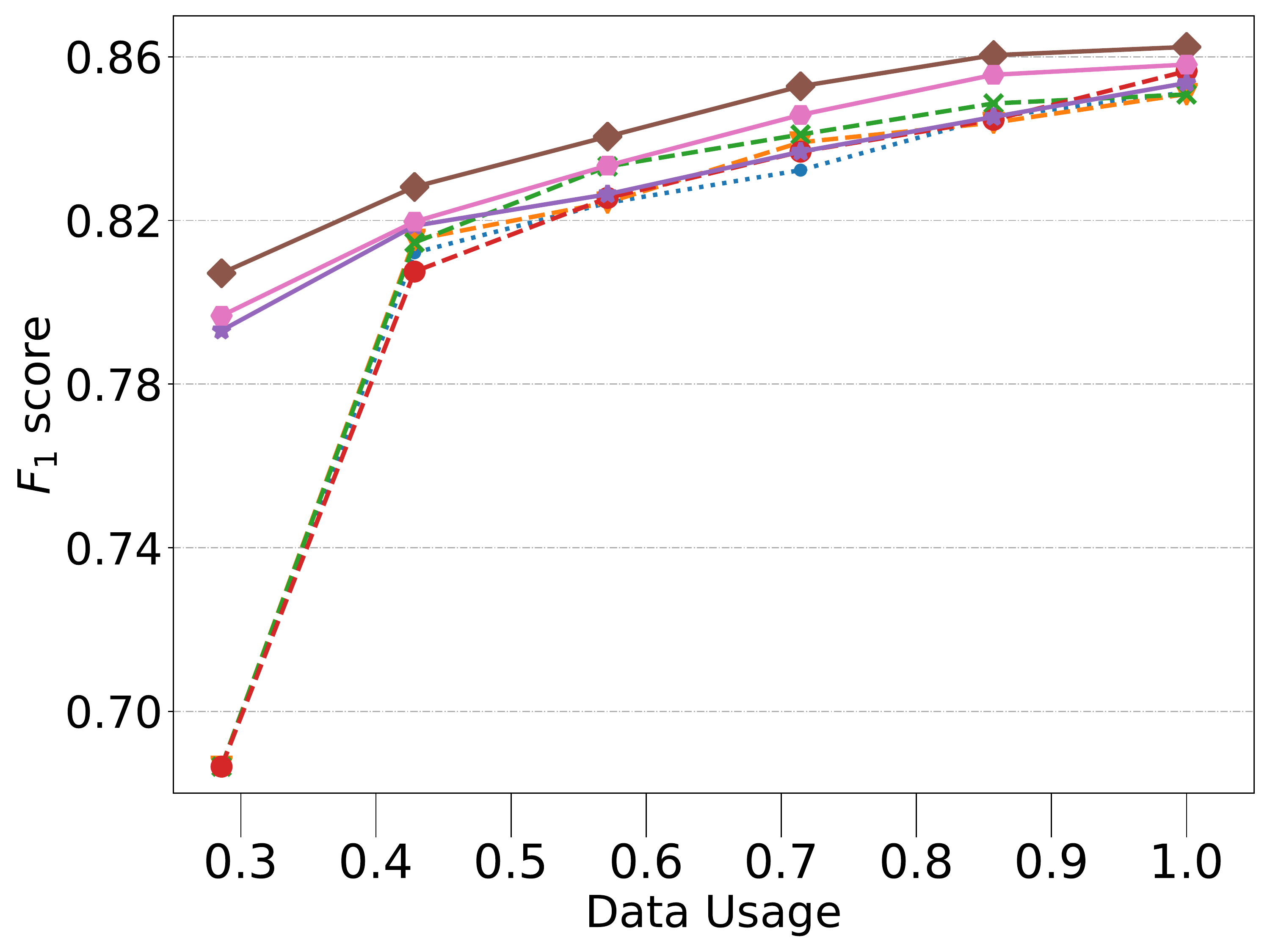}
            \vspace{-0.2in}
            \label{fig:conll}
        }\hspace{-3mm}
        \subfigure[ACE05 (14k labeled data)]{
            \includegraphics[width=0.322\textwidth]{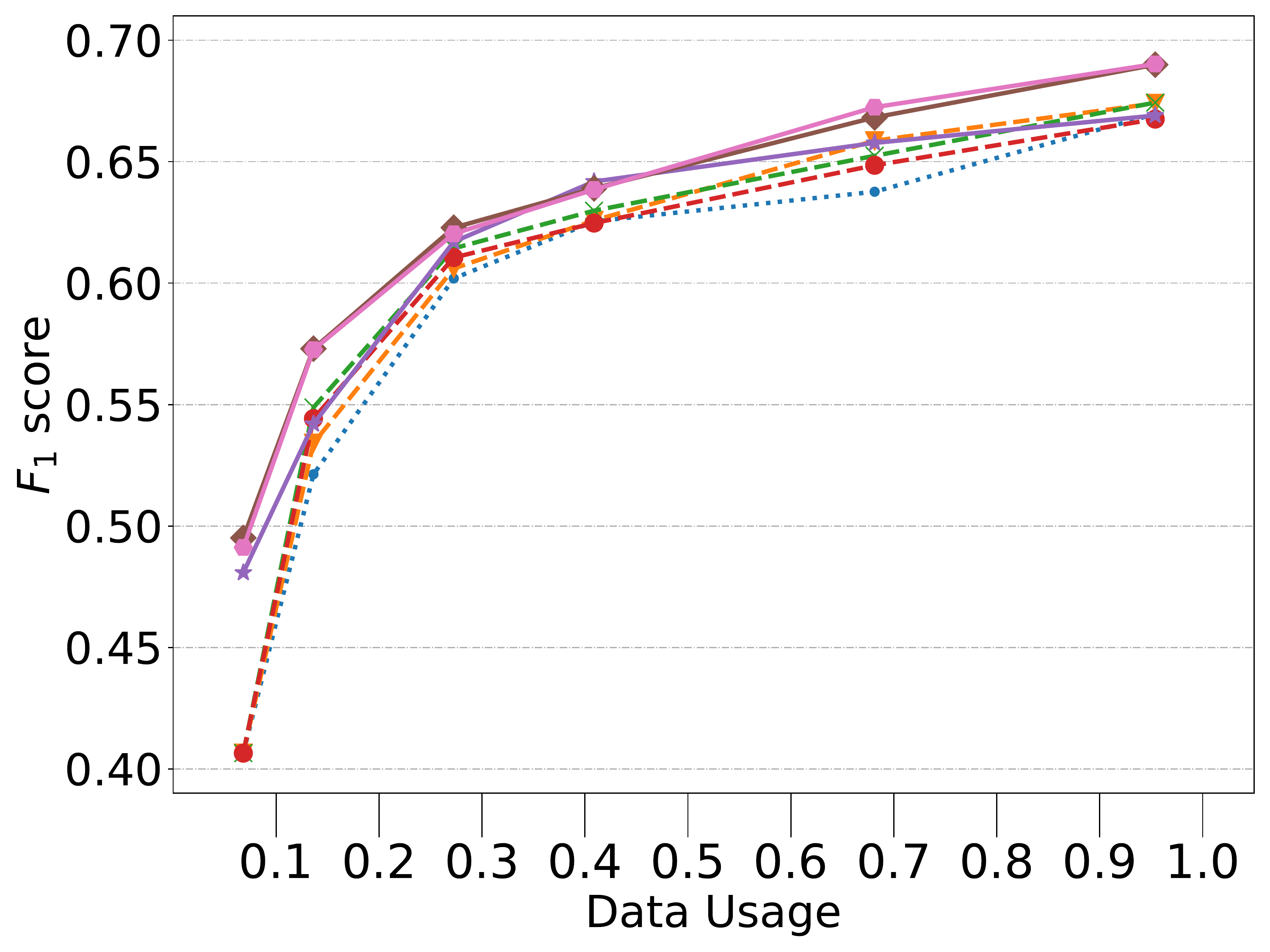}
            \vspace{-0.2in}
            \label{fig:ace}
        }\hspace{-3mm}
         \subfigure[WebPage (385 labeled data)]{
            \includegraphics[width=0.322\textwidth]{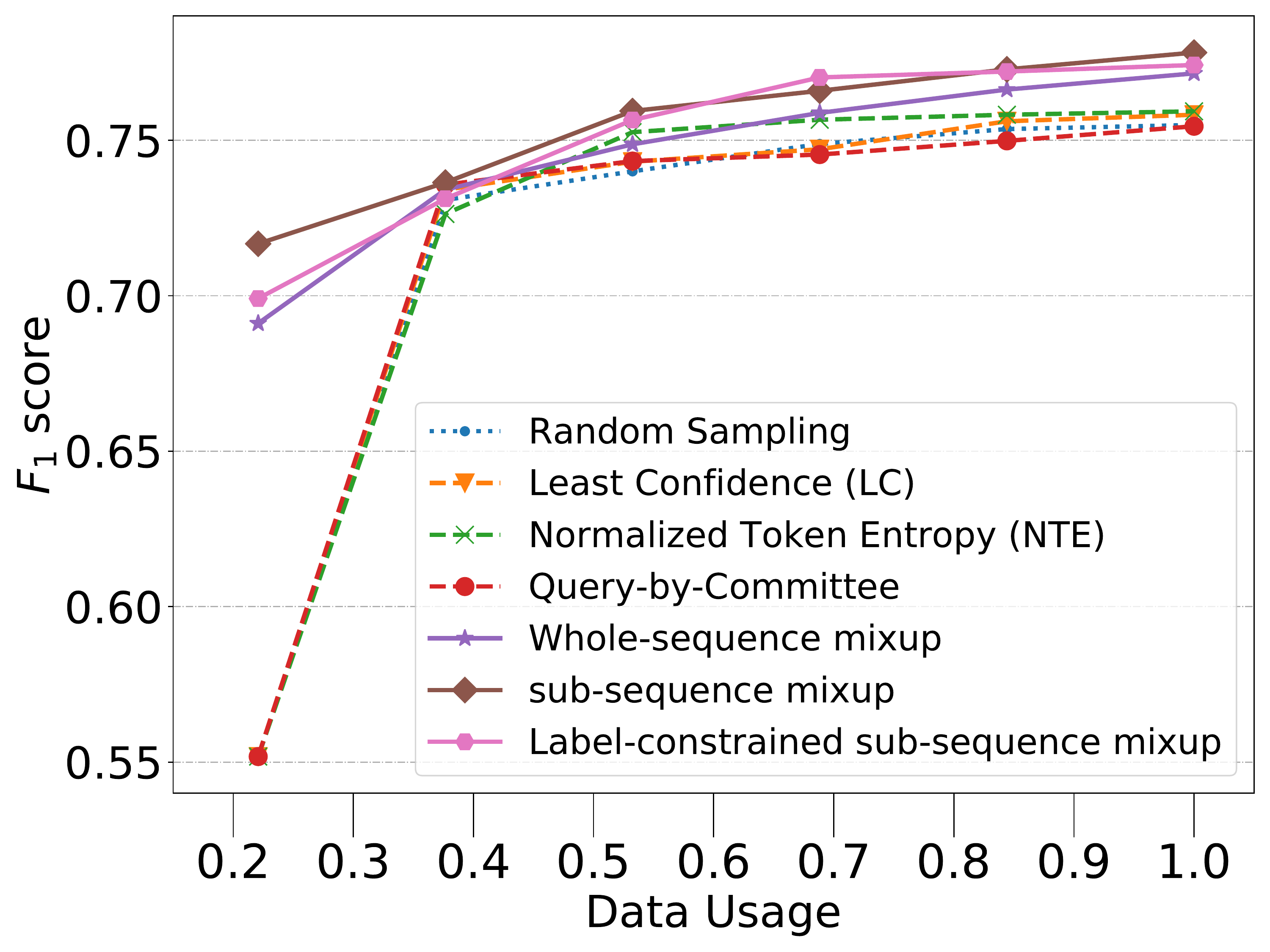}
            \vspace{-0.2in}
            \label{fig:webpage}
        }
       \vspace{-0.07in}
        \caption{The $F_1$ score of test set in terms of data usage on CoNLL-03, ACE05 and WebPage.}
        \label{fig:main_result}
	\vspace{-0.07in}
\end{figure*}
\begin{figure}
    \centering
    \includegraphics[scale=0.22]{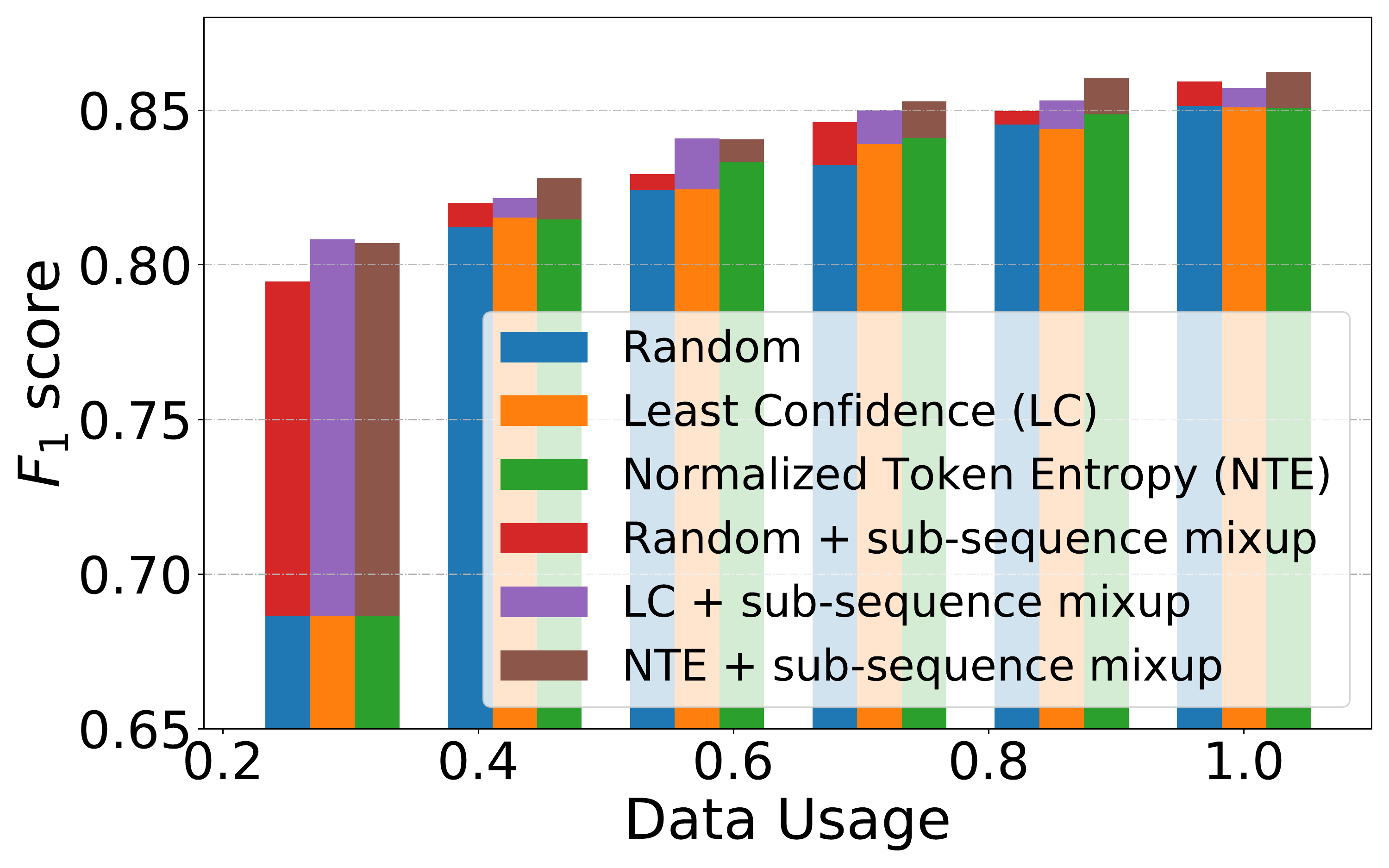}
    \vspace{-2mm}
    \caption{The improvements to various active learning approaches provided by {\ours}.}
    \label{fig:active_seqmix}
    \vspace{-4mm}
\end{figure}

The main results are presented in Figure \ref{fig:main_result}, where we use NTE sampling as the default active learning policy.
From the result, it is clear that our method achieves the best performance consistently at each data usage percentile for all three datasets. The best {\ours} method (sub-sequence mixup with NTE sampling) outperforms the strongest active learning baselines by $2.95\%$ on CoNLL-03, $2.27\%$ on ACE05 and $3.75\%$ on WebPage in terms of $F_1$ score on average.
Moreover, the augmentation advantage is especially prominent for the seed set initialization stage where we only have a very limited number of labeled data. Through the augmentation, we improve the model performance from $68.65\%$ to $80.71\%$, where the seed set is 200 labeled sequences and the augmentation provides extra 40 data points for CoNLL-03. The improvement is also significant on ACE05 ($40.65\%$ to $49.51\%$), and WebPage ($55.18\%$ to $71.67\%$), which indicates that our {\ours} can largely resolve the label scarcity issue in low-resource scenarios.

We also perform
statistical significance tests for the above results. We use Wilcoxon Signed Rank Test \cite{wilcoxon1992individual}, a non-parametric alternative to the paired t-test. This significance test fits our task as F-score is generally assumed to be not normally distributed \cite{dror2018hitchhiker}, and non-parametric significance tests should be used in such a case. The results show that sub-sequence mixup and label-constrained sub-sequence mixup can provide a statistical significance (the confidence level $\alpha = 0.05$ and the number of data points $N = 6$) for all the comparisons with active learning baselines on used datasets. The whole-sequence mixup passes the statistical significance test with $\alpha = 0.1$ and $N = 6$ on CoNLL-03 and WebPage, but fails on ACE05.

Among all the three {\ours} variants, sub-sequence mixup gives the overall best performance (label-constrained sub-sequence mixup achieves very close performance with sub-sequence mixup on ACE05 dataset), but whole-sequence mixup does not yield a consistent improvement to the original active learning method. 
This is because the whole-sequence mixup may generate semantically poor new sequences.
Instead, the sub-sequence-level process reserves the original context information between the sub-sequence and the other parts of the whole sequence. Meanwhile, the updated sub-sequences inherit the original local informativeness, and introduce linguistic diversity to enhance the model's generalization ability.

To justify that {\ours} can provide improvement to the active learning framework with various query policies, we employ different query policies with {\ours} augmentation under the same experiment setting as Figure \ref{fig:conll}.
From Figure \ref{fig:active_seqmix}, we find that there is a consistent performance improvement when employing {\ours} with different query policies. As {\ours} achieves $\{2.46\%, 2.85\%, 2.94\%\}$ performance gain  for random sampling, LC sampling and NTE sampling respectively.



\subsection{Effect of Discriminator}
\label{ablation_study}
\begin{table}[t]
\setlength{\tabcolsep}{1mm}
\centering
\begin{small}
\begin{tabular}{c|cccccc}\hline
 Data Usage & 200    & 300    & 400    & 500    & 600    & 700    \\\hline
(0, $+\infty$)      & \textbf{81.15} & 82.32 & 82.74 & 83.66 & 83.79 & 85.05 \\
(0, 2000)   & 80.20 & 82.24 & 83.21 & 83.67 & 83.90 & 85.11 \\
(0, 1000)   & 80.13 & 81.86 & 83.58 & 84.22 & 84.81 & 85.16 \\
(0, 500)    & 80.71 & \textbf{82.82} & \textbf{84.05} & \textbf{85.28} & \textbf{86.04} & \textbf{86.24} \\
\hline
\end{tabular}
\end{small}
\caption{\label{score_range} The $F_1(\%)$ of sub-sequence mixup with NTE sampling in different discriminator score range, evaluated on CoNLL-03 with 700 data.}
\vspace{-4mm}
\end{table}

\begin{table*}[h]
\setlength{\tabcolsep}{1mm}
\centering
\begin{small}
\begin{tabular}{c|cccccc|c}\hline
\textbf{Data Usage} & 200    & 300    & 400    & 500    & 600    & 700 & \textbf{Average}   \\\hline
$r=0.2$      &\textbf{80.22 (+0.76)} & 82.23(+0.43) &\textbf{83.61 (+0.61)}  & 84.62 (+0.53) & 85.16 (+0.10)& 85.22 (-0.11) & \textbf{+ 0.39}\\
$r=0.4$  &79.71 (+0.25) &\textbf{82.48(+0.68)} &82.66 (-0.34)  & 83.46 (-0.63) & 84.79 (-0.27)& 85.24 (-0.09)& - 0.07 \\
$r=0.6$   &79.40 (-0.06)& 82.07(+0.27)    &83.34 (+0.34)  & \textbf{84.75 (+0.66)} & \textbf{85.43 (+0.37)} & \textbf{85.50 (+0.17)}& + 0.29 \\
$r=0.8$    &79.48 (+0.02)& 81.63(-0.17)    &82.80 (-0.20)  & 83.29 (-0.80) & 84.54 (-0.52) & 85.32 (-0.01)& - 0.28\\
$r=1.0$  &78.51 (-0.95)& 80.58(-1.22)      &82.59 (-0.41)  & 84.31 (+0.22) & 85.36 (+0.30) & 85.37 (+0.04)& - 0.34\\
\hline
\end{tabular}
\end{small}
\caption{The $F_1$ score with variant augment rate $r$. The value in the parentheses is the difference with the average $F_1$ for corresponding data usage. The last column presents the average $F_1$  difference for each learning rate $r$.}
 \label{aug_rate}
\end{table*}
To verify the effectiveness of the discriminator, we conduct the ablation study on a subset of CoNLL-03 with 700 labeled sequences. We use sub-sequence mixup with NTE sampling as the backbone and change the perplexity score range of the discriminator. We start from the seed set with 200 labeled data, then actively query 100 data in each learning round and repeat 5 rounds in total.

The result in Table \ref{score_range} demonstrates the
discriminator provides a stable improvement for the last
four data usage percentiles, and the discriminator with score range $(0, 500)$
can boost the model by $1.07\%$ $F_1$ score, averaged by all the data usage
percentiles. The comparison between 3 different score thresholds demonstrates
the lower the perplexity, the better the generation quality. As a result, the final   $F_1$ score becomes higher with the better generated tokens. Actually, we can further narrow
down the score range to get more performance improvement in return, but the too
strict constraints will slow down the generation in practice and reduce the
number of generated samples.

\subsection{Parameter Study}
\label{param_study}
In this subsection, we study the effect of several key parameters.


\paragraph{Augment rate $r$.} 
We vary the augment rate $r = \frac{|\mathcal{L}^*|}{|\psi(\mathcal{U},K,\gamma(\cdot))|}$ in $\{0.2, 0.4, 0.6, 0.8, 1.0\}$ and keep the number of initial data usage same to investigate the effect of augment rate for data augmentation. 
Table \ref{aug_rate} shows that $r \leq 0.6$ can provide better $F_1$ improvement. The model with $r = 0.2$ surpasses the model with $r = 1.0$ by $0.73\%$, evaluated by the average $F_1$ score for all the data usage percentiles. This result indicates that the model appreciates moderate augmentation more. However, the performance variance based on the augment rate is not prominent compared to the improvement provided by {\ours} to the active learning framework.

\paragraph{Valid tag density $\eta_0$.}
\begin{figure}[t]
    \vspace{-0.05in}
        \centering
        \subfigure[$F_1$ with several combination of $s$ and $n$]{
            \includegraphics[width=0.231\textwidth]{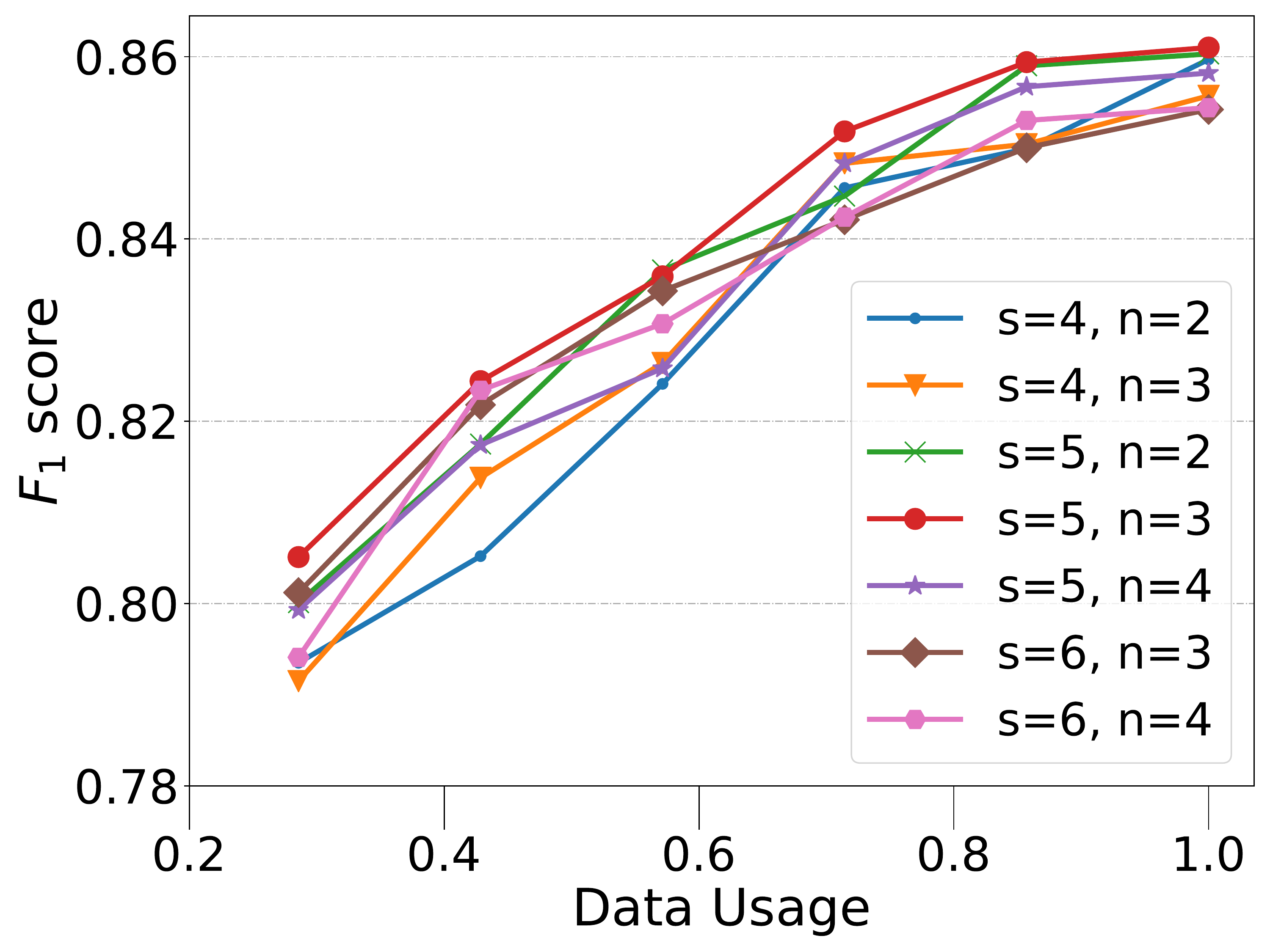}
            \vspace{-0.2in}
            \label{fig:vtd}
        }\hspace{-3.7mm}
        \subfigure[$F_1$ with different $\alpha$]{
            \includegraphics[width=0.231\textwidth]{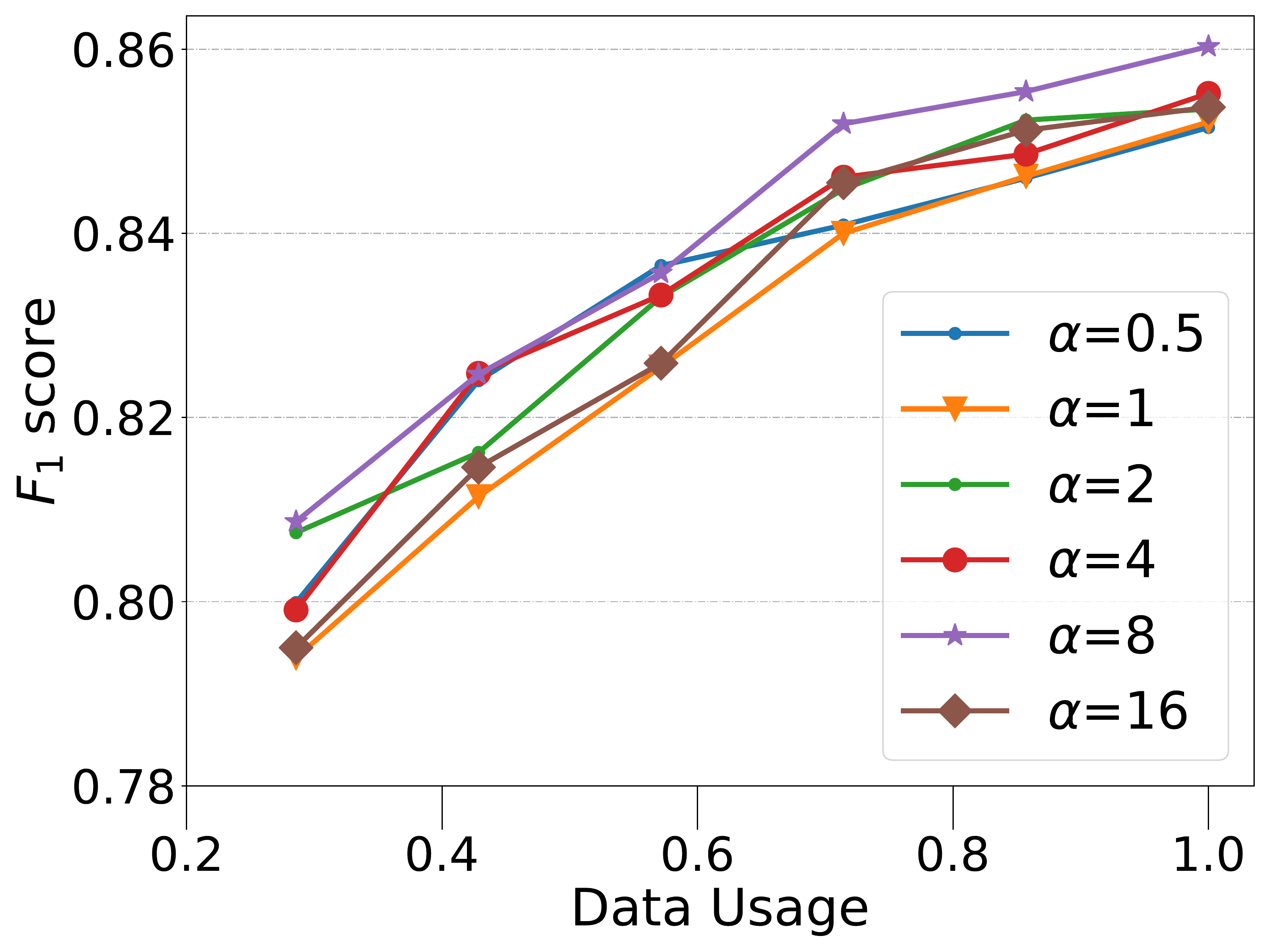}
            \vspace{-0.2in}
            \label{fig:alpha}
        }
       \vspace{-0.15in}
        \caption{Parameter Search for {\ours}}
        \label{fig:para}
	\vspace{-0.15in}
\end{figure}
We search the valid tag density $\eta_0$ as Section \ref{mixup_emb} defined by varying the sub-sequence window length $s$ and the required number of valid tag $n$ within the window.   The results in Figure \ref{fig:vtd} illustrate the combination $(s=5, n=3)$ outperforms other settings. When $s$ is too small, the window usually truncates the continuous clause, thus cutting off the local syntax or semantic information. When $s$ is too large, sub-sequence mixup tends to behave like whole-sequence {\ours}, where the too long sub-sequence generation can hardly maintain the rationality of syntax and semantics as before. The high $\eta_0$ with long window length may result in an insufficient amount of eligible parent sequences. Actually, even with a moderate augment rate $\alpha = 0.2$, the combination $(s=6, n=5)$ has been unable to provide enough generation.

\paragraph{Mixing parameter $\alpha$.}
We show the performance with different $\alpha$ in Figure \ref{fig:alpha}. The parameter $\alpha$ decides the distribution $\lambda \sim \textit{Beta}(\alpha, \alpha)$, and the coefficient $\lambda$ directly involved the mixing of tokens and labels. Among the values $\{0.5, 1, 2, 4, 8, 16\}$, we observed $\alpha = 8$ presents the best performance. It outperforms the second-best parameter setting $0.49\%$ by average. From the perspective of Beta distribution, larger $\alpha$ will make the sampled $\lambda$ more concentrated around 0.5, which assigns more balance weights to the parent samples to be mixed. In this way, the interpolation produces encoded token with further distance to both the parent samples, thus introduces a more diverse generation.

\subsection{Case Study}
\begin{figure*}[h]
  \centering
  \includegraphics[scale = 0.3]{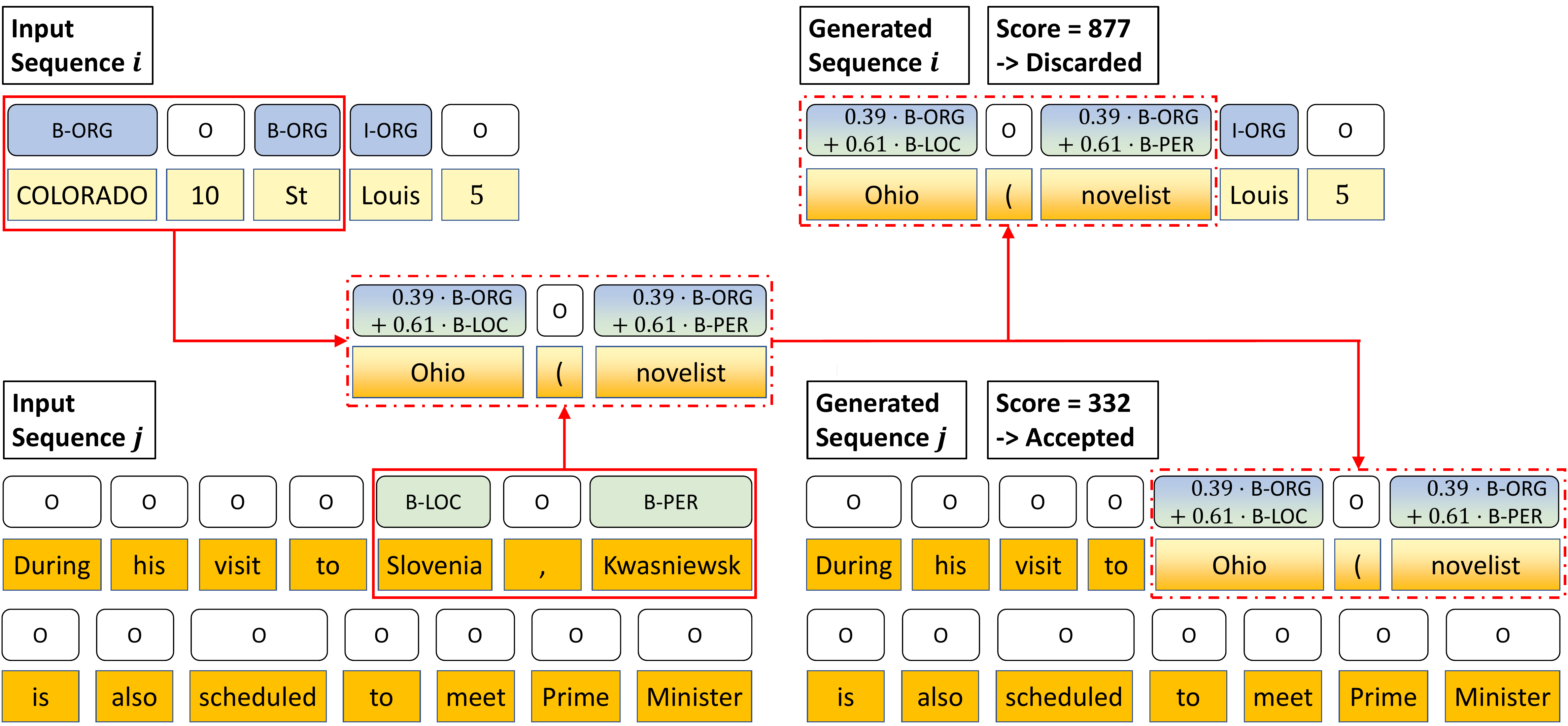}
  \caption{A generation case of sub-sequence mixup.}
  \label{fig:case_study}
  \vspace{-3mm}
\end{figure*}

Figure \ref{fig:case_study} presents a generation example via sub-sequence mixup.
For the convenience of presentation, we set the length of sub-sequence $s=3$
and the valid label density threshold $\eta_0 = \frac{2}{3}$. The two input sequences got
paired for their eligible sub-sequences ``\texttt{COLORADO 10 St}'' and ``\texttt{Slovenia , Kwasniewski}''. The sub-sequences are mixed by $\lambda = 0.39$ in this case, which is sampled from
\textit{Beta}$(\alpha, \alpha)$.
Then the generated sub-sequence ``\texttt{Ohio ( novelist}'' replaces the original parts in the two input sequences. Among the generated tokens, ``\texttt{Ohio}'' inherits the label \texttt{B-ORG} from ``\texttt{COLORADO}'' and the label \texttt{B-LOC} from ``\texttt{Slovenia}'', and the distribution \textit{Beta}$(\alpha, \alpha)$ assigns the two labels with weights $\lambda = 0.39$ and $(1-\lambda) = 0.61$. The open parenthesis is produced by the mixing of
a digit and a punctuation mark, and keeps the label \texttt{O} shared by its parents. Similarly, the token ``\texttt{novelist}'' generated by ``\texttt{St}'' and ``\texttt{Kwasniewski}'' gets a mixed label from \texttt{B-ORG} and \texttt{B-PER}.

The discriminator then evaluates the two generated sequences. The generated sequence $i$ is not reasonable enough intuitively, and its perplexity score $877$ exceeds the threshold, so it is not  added into the training set. The generated sequence $j$ retains the original syntax and semantic structure much better. Although the open parenthesis seems strange, it plays a role as the comma in the original sequence to separate two clauses. This generation behaves closely to a normal sequence and earns $332$ perplexity score, which permits its incorporation into the training set.

\section{Related Work}
\paragraph{Active Sequence Labeling}
Sequence labeling has been studied extensively for different NLP problems. Different neural architectures has been proposed
\cite{huang2015bidirectional, lample2016neural,peters2018deep,akbik2018contextual} in recent years, which have achieved
state-of-the-art performance in a number of sequence labeling tasks. However,
these neural models usually require exhaustive human efforts for generating labels for each token, and may not perform well in low-resource settings. 
To improve the performance of low-resource sequence labeling, several approaches have been applied including using semi-supervised methods~\cite{cvt,seqvat}, external weak supervision~\cite{lison2020named,bond,ren2020denoise,zhang2019how,yu2020steam} and active learning~\cite{shen2017deep,hazra2019active,liu2018learning,fang2017learning,gao2019active}. 
In this study, we mainly focus on
active learning approaches which select samples
based on the query policy design. So far, various uncertainty-based \cite{scheffer2001active, culotta2005reducing, kim2006mmr} and committee-based approaches \cite{dagan1995committee} have been proposed for improving the sample efficiency. 
More recently, \citet{shen2017deep,hazra2019active,liu2018learning,fang2017learning} further improve the aforementioned active learning approaches to
improve the sampling diversity as well as the model's generalization ability on low-resource scenarios. These works mainly claim the sample efficiency
provided by the active learning approach but do not study data augmentation for active sequence labeling.

\paragraph{Interpolation-based Regularizations}
Mixup implements interpolation in the input space to regularize models \cite{zhang2017mixup}. Recently, the Mixup variants \cite{verma2018manifold, summers2019improved, guo2019mixup} turn to perform interpolation in the hidden space to capture higher-level information. 
\citet{guo2019augmenting,chen2020mixtext} apply hidden-space Mixup for text classification. These works, however, have not explored how to perform
mixup for sequences with token-level labels, nor
do they consider the quality of the mixed-up samples.


\paragraph{Text Augmentation}
Our work is also related to text data augmentation. \citet{zhang2015character,wei2019eda} utilize heuristic approaches including synonym replancement, random insertion, swap and deletion for text augmentation, \citet{kafle2017data, silfverberg2017data} employ heuristic rules based on specific task, \citet{hu2017toward} propose to augment text data in an encoder-decoder manner. 
Very recently, \cite{lambada20,kobayashi2018contextual} harness the power of pre-trained language models and augmenting the text data based on contextual patterns. Although these methods can augment the training set and improve the performance of text classification model, they fail to  \emph{generate sequences and labels simultaneously}, thus cannot be adapted to our problem where token-level labels are required during training.
Instead, in our study, we propose a new framework {\ours} for data augmentation to facilitate sequence labeling task. Our method can generate token-level labels and preserve the semantic information in the augmented sentences. Moreover, it can be naturally combined with existing active learning approaches and further promote the performance.
\section{Conclusion}


We proposed a simple data augmentation method {\ours} to enhance active sequence labeling. By performing sequence mixup in the latent space, {\ours}
improves data diversity during active learning, while being able to generate plausible augmented sequences. This method is generic to different active
learning policies and various sequence labeling tasks. Our experiments demonstrate that {\ours} can improve active learning baselines consistently for
NER and event detection tasks; and its benefits are especially prominent in low-data regimes. For future research, it is interesting to enhance
{\ours} with language models during the mixup process, and harness external knowledge for further improving diversity and plausibility.



\bibliography{emnlp2020}
\bibliographystyle{acl_natbib}

\clearpage

\appendix
\section{Information for Dataset}
\label{app:dataset}
\subsection{Dataset Collection}
Here we list the link to datasets used in our experiments.
\begin{itemize}
    \item  \textbf{CoNLL-03}: \url{https://github.com/synalp/NER/tree/master/corpus/CoNLL-2003}.
    \item \textbf{ACE05}: We are unable to provide the downloadable version due to it is not public. This corpus can be applied through the website of LDC: \url{https://www.ldc.upenn.edu/collaborations/past-projects/ace}.
    \item \textbf{Webpage}: Please refer the link in the paper \cite{ratinov2009design}.
\end{itemize}

\begin{table*}[tb]
\begin{small}
	\begin{center}
		\begin{tabular}{c|c|c|c|c|c|c}
			\toprule 
			\bf Dataset & \bf \# of Entity Types& \bf \# of Seed Set & \bf Sampling Rounds & \bf \# of Each Round Samples & \bf  \# of Dev & \bf  \# of Test \\ \midrule 
			 CoNLL-03 & 4 & 200 & 5 & 100 & 3250 & 3453 \\ \hline 
			 ACE05 & 29 & 1k & 5 & \{1k, 2k, 2k, 4k, 4k\} & 873& 711 \\\hline 
			 Webpage & 4 & 85 & 5 & 60 & 99 & 135 \\ \bottomrule
		\end{tabular}
	\end{center}
	\caption{The information for  benchmarks in our experiments.}
	\label{tab:dataset}
\end{small}
\end{table*}

\subsection{Dataset Split}
All the mentioned dataset has been split into \textbf{train/validate/test} set in the released version.
We keep consistent with the validation set and the test set in our experiment. For the active learning paradigm, we split the training set as Table \ref{tab:dataset}. The active learners are initialized on the seed set, then they implement 5 active learning rounds.

\section{Baseline Settings}

For the baselines, we take random sampling and 3 active learning approaches -- LC sampling, NTE sampling, and QBC sampling as Section \ref{sampling_policy}.

\section{Implementation Details of SeqMix}
We implement bert-base-cased as the underlying model for the NER task and bert-base-multilingual-cased as the underlying model for the event detection task. We use the model from Huggingface Transformer codebase\footnote{\url{https://github.com/huggingface/transformers}}, and the repository\footnote{\url{https://github.com/kamalkraj/BERT-NER}} to fine-tune our model for sequence labeling task.
\subsection{Number of Parameters}
In our model, we use \textbf{bert-base-cased} and \textbf{bert-base-multilingual-cased} both of them occupy 12-layer, 768-hidden, 12-heads with 110M parameters. 

\subsection{Adapting BERT for sequence labeling task}

To fine-tune on sequence labeling tasks, a dropout layer ($p=0.1$) and a linear (token-level) classification layer is built upon the pre-trained model.

\subsection{SeqMix Details}
In Section \ref{mixup_emb}, we construct a table of tokens $\mathcal{W}$ and their corresponding contextual embedding $\mathcal{E}$. For our underlying BERT model, we use the vocabulary provided by the tokenizer to build up $\mathcal{W}$, and the embedding initialized on the training set as $\mathcal{E}$.

We also need to construct a special token collection to exclude some generation in the process of sequence mixing. For example, BERT places token \texttt{[CLS]} and \texttt{[SEP]} at the starting position and the ending position for sentence, and pad the inputs with \texttt{[PAD]}. We exclude these disturbing tokens and the parent tokens.

\subsection{Parameter Settings}
The key parameters setting in our framework are stated here:
(1) The number of active learning round is 5 for all the three datasets, but the size of seed set and the number of samples in each round differs from the dataset. We list the specific numbers as Table \ref{tab:dataset}.
(2) The sub-sequence window length $s$ and the valid label density threshold $\eta_0$ vary from the datasets.
For CoNLL-03, $s = 5$, $\eta_0 = 0.6$;
for ACE05, $s = 5$, $\eta_0 = 0.2$;
for WebPage, $s  = 4$, $\eta_0 = 0.5$.
(3) We set $\alpha=8$ for the \textit{Beta} distribution.
(4) The discriminator score range is set as $(0, 500)$ for all the datasets.
(5) For BERT configuration, we choose 5e-5 for learning rate, 128 for padding length, 32 for batch size, 0.1 for dropout rate, 1e-8 for $\epsilon$ in Adam. At each data usage point, we train the model for 10 Epochs.
(6) We set $\mathcal{C} = 3$ for the QBC query policy.

\section{Details of Experiments}
We take following criteria to evaluate the sequence labeling task. A named entity is correct only if it is an exact match of the corresponding entity in the data file. An event trigger is correct only if the span and type match with golden labels. Based on the above metric, we evaluate $F_1$ score in our experiments.

\subsection{Performance on Development Set}

\begin{table}[h]
\setlength{\tabcolsep}{0.75mm}
\centering
\begin{small}
\begin{tabular}{c|cccccc}\hline
\textbf{Data Usage} & 200    & 300    & 400    & 500    & 600    & 700 \\\hline
Random Sampling & 69.03& 83.28& 84.93& 85.50& 85.79& 86.62\\
LC Sampling &69.03& 83.78& 84.55& 85.88& 86.04& 86.73  \\
NTE Sampling & 69.03& 83.60& 85.00& 85.47& 86.19& 86.83\\
QBC Sampling &69.03 &83.33 &84.52 & 85.30& 86.27& 86.60 \\
Sub-sequence mixup & 81.69& 85.28& 85.95& 86.52& 87.07& 87.44\\
\hline
\end{tabular}
\end{small}
\caption{Validation $F_1$ of CoNLL-03}
 \label{dev_conll}
\end{table}
\begin{table}[h]
\setlength{\tabcolsep}{0.75mm}
\centering
\begin{small}
\begin{tabular}{c|cccccc}\hline
\textbf{Data Usage} & 1000    & 2000    & 4000    & 6000    & 10000  & 14000 \\\hline
Random Sampling & 48.16& 59.10& 63.13& 64.95& 66.23 & 67.12\\
LC Sampling & 48.16 & 59.33& 63.22& 65.04& 66.24& 66.92 \\
NTE Sampling & 48.16& 59.72& 63.17& 65.53& 66.78& 67.24 \\
QBC Sampling & 48.16& 59.01& 62.79& 64.89& 66.20& 66.91  \\
Sub-sequence mixup  & 56.51& 61.62& 63.65& 65.83 & 67.54& 67.98\\
\hline
\end{tabular}
\end{small}
\caption{Validation $F_1$ of ACE05}
 \label{dev_ace}
\end{table}
\begin{table}[h]
\setlength{\tabcolsep}{0.75mm}
\centering
\begin{small}
\begin{tabular}{c|cccccc}\hline
\textbf{Data Usage} & 85    & 145    & 205   & 265    & 325    &385 \\\hline
Random Sampling &0& 27.52& 34.41& 34.83& 37.93& 35.73\\
LC Sampling &0& 28.84& 32.88& 34.22& 38.78& 38.11  \\
NTE Sampling & 0& 22.44& 34.81& 33.74& 36.59& 38.27\\
QBC Sampling &0 & 23.88& 32.18& 34.17&36.56& 35.66\\
Sub-sequence mixup  &14.35& 33.74& 34.70& 36.22& 39.74& 38.25 \\
\hline
\end{tabular}
\end{small}
\caption{\label{augment_rate} Validation $F_1$ of WebPage}
 \label{dev_webpage}
\end{table}
Table \ref{dev_conll} to Table \ref{dev_webpage} shows the model performance on the validation set.
The data usage in these tables refers to the number of labeled data, excluding the augmentation data.
Sub-sequence mixup is trained with $(1+\alpha)$ times data, where the $\alpha$ denotes the augment rate.
Note that WebPage is a very limited dataset, there is a big difference between the performance on the validation set and the test set.
We average each experiment by 5 times.

\subsection{Computing Infrastructure}
We implement our system on \textit{Ubuntu 18.04.3 LTS} system. We run our experiments on an Intel(R) Xeon(R) CPU @ 2.30GHz and NVIDIA Tesla P100-PCIe with 16 GB HBM2 memory. The NVIDIA-SMI version is 418.67 and the CUDA version is 10.1.

\subsection{Average Runtime}
For the 5-round active learning with SeqMix augmentation, our program runs about
500 seconds for WebPage dataset, 1700 seconds for the CoNLL slicing dataset, and 3.5 hours for ACE 2005.
If the QBC query policy used, all the runtime will be multiplied about 3 times.

\subsection{Hyper parameter Search}
For the discriminator score range, we first examine the perplexity score distribution of the CoNLL training set. Then determine an approximate score range $(0, 2000)$ first. We linearly split score ranges below 2000 to conduct parameter study and report the representative ranges in Section \ref{ablation_study}. Given the consideration to the generation speed and the augment rate setting, we finally choose $500$ as the upper limit rather than a too narrow score range setting.

For the mixing coefficient $\lambda$, we follow \cite{zhang2017mixup} to sample it from \textit{Beta}$(\alpha, \alpha)$ and explore $\alpha$ ranging from $[0.5, 16]$. We present this parameter study in Section \ref{param_study}. The result shows different $\alpha$ did not influence the augmentation performance much.

For the augment rate and the valid tag density, we also have introduced the parameter study in Section \ref{param_study}.

\end{document}